\documentclass[10pt,twocolumn,letterpaper]{article}

\usepackage[pagenumbers]{iccv} 
\usepackage{multirow}
\usepackage{algorithmic}
\usepackage{algorithm}
\usepackage[table]{xcolor} %

\newcommand{\colorfirst}{255, 153, 153}
\newcommand{\colorsecond}{255, 204, 153}
\newcommand{\colorthird}{255, 255, 153}
\definecolor{colorfirst}{RGB}{255, 153, 153}
\definecolor{colorsecond}{RGB}{255, 204, 153}
\definecolor{colorthird}{RGB}{255, 255, 153}

\definecolor{colorfirst}{RGB}{255, 153, 153}
\definecolor{colorsecond}{RGB}{255, 204, 153}
\definecolor{colorthird}{RGB}{255, 255, 153}

\newcommand{\teaser}{
\vspace{-2em}
\centering
\includegraphics[width=\textwidth,trim=0em 0em 0em 0em,
clip]{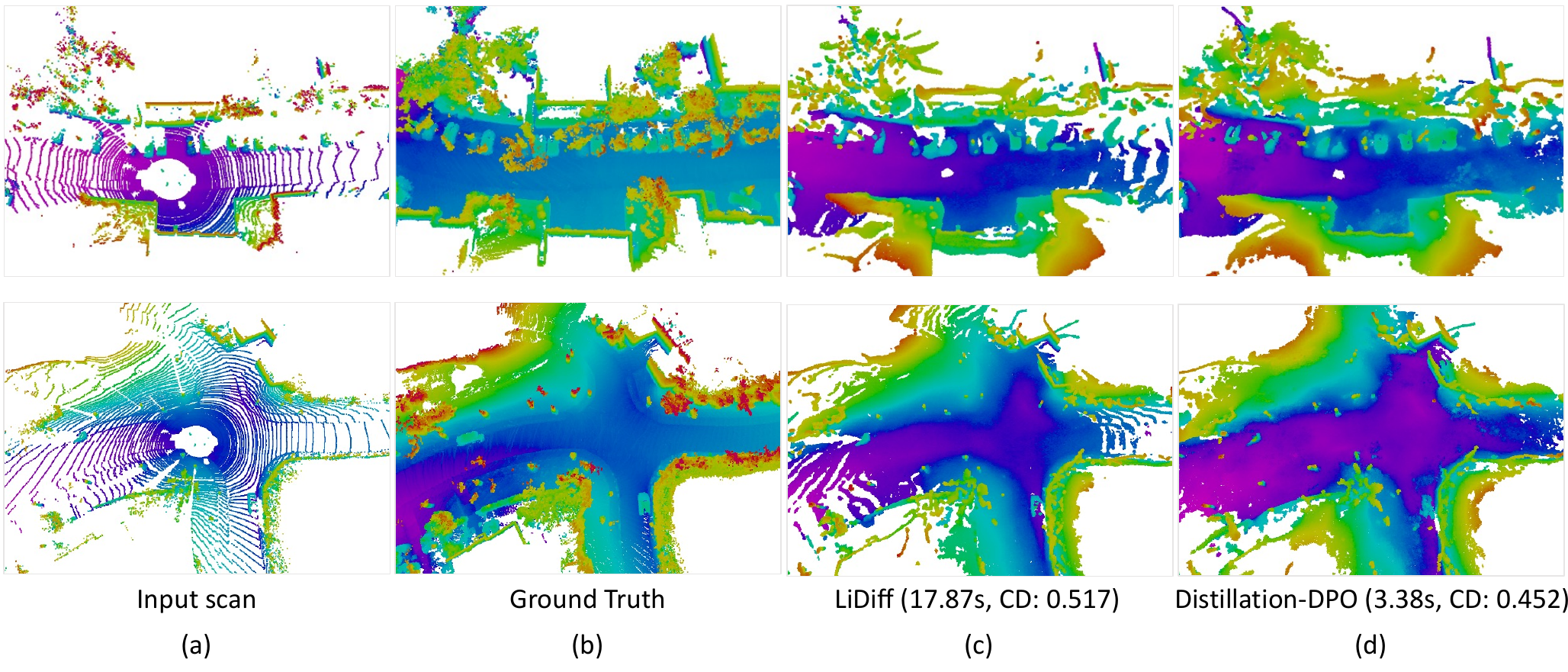}
\vspace{-1.1em}
\captionof{figure}{
An example demonstration of Distillation-DPO for LiDAR scene completion on SemanticKITTI dataset. (a) The input sparse LiDAR scan. (b) The corresponding ground truth scene. (c) Completion results of the existing state-of-the-art (SOTA) model, LiDiff~\cite{LiDiff}. (d) Completion results of the proposed Distillation-DPO. Compared to LiDiff, Distillation-DPO can complete a scene more than 5 times faster while achieving higher completion quality (lower Chamfer Distance).
}
\label{fig:teaser}
\vspace{1em}
}

\definecolor{iccvblue}{rgb}{0.21,0.49,0.74}
\usepackage[pagebackref,breaklinks,colorlinks,allcolors=iccvblue]{hyperref}

\def\paperID{4245} 
\def\confName{ICCV}
\def\confYear{2025}

\title{Diffusion Distillation With Direct Preference Optimization For Efficient 3D LiDAR Scene Completion}

\author{An Zhao$^1$
\and Shengyuan Zhang$^1$ \and Ling Yang$^2$ \and Zejian Li$^1$ \and Jiale Wu$^1$ \and Haoran Xu$^3$ \and AnYang Wei$^3$ \and Perry Pengyun GU$^3$ \and Lingyun Sun$^1$ \\
{\small $^1$ Zhejiang University \quad \small $^2$ Peking University \quad \small $^3$ Zhejiang Green Zhixing Technology co., ltd} \\
{\tt\small $^1$ \{zhangshengyuan,zhaoan040113,zejianlee,ialewu2022,sunly\}@zju.edu.cn} \\
{\tt\small$^2$ \{yangling0818\}@163.com } \\
{\tt\small$^3$ \{Haoran.Xu5,weianyang,gupengyun\}@geely.com}
}

\begin{document}
\twocolumn[{
\maketitle
\teaser
}]
\begin{abstract}
The application of diffusion models in 3D LiDAR scene completion is limited due to diffusion's slow sampling speed. 
Score distillation accelerates diffusion sampling but with performance degradation, while post-training with direct policy optimization (DPO) boosts performance using preference data.
This paper proposes Distillation-DPO, a novel diffusion distillation framework for LiDAR scene completion with preference aligment.
First, the student model generates paired completion scenes with different initial noises.
Second, using LiDAR scene evaluation metrics as preference, we construct winning and losing sample pairs. 
Such construction is reasonable, since most LiDAR scene metrics are informative but non-differentiable to be optimized directly.
Third, Distillation-DPO optimizes the student model by exploiting the difference in score functions between the teacher and student models on the paired completion scenes.
Such procedure is repeated until convergence.
Extensive experiments demonstrate that, compared to state-of-the-art LiDAR scene completion diffusion models, Distillation-DPO achieves higher-quality scene completion while accelerating the completion speed by more than 5-fold.
Our method is the first to explore adopting preference learning in distillation to the best of our knowledge and provide insights into preference-aligned distillation.
Our code is public available on \url{https://github.com/happyw1nd/DistillationDPO}.

\end{abstract}    
\section{Introduction}
\label{sec:intro}



Recently, the diffusion model has gradually been utilized for LiDAR scene completion due to the outstanding performance in image super-resolution~\cite{image_super1,image_super2} and video synthesis~\cite{video1,video2}. However, since LiDAR point cloud completion requires high-precision set reconstruction and high-quality completion of missing points, diffusion models often need to sacrifice sampling time to achieve high-quality completion results. Thus, despite the potential of diffusion models in this domain, the slow sampling speed limits their practicality in real-world applications. 

As an effective distillation method for the diffusion model, the effectiveness of score distillation has been well established~\cite{Diff-Instruct,DMD,DMD2}, which provides an effective pathway for accelerating LiDAR scene completion diffusion models. However, score distillation inevitably leads to information loss and a quality decline in the completed scene during the sampling acceleration process.

Reward models provide a potential way to mitigate the performance degradation caused by distillation. 
The reward model learns human preferences to predict the rating of generated samples, while existing methods primarily enhance generation quality by maximizing the rating predicted by the reward model~\cite{itercomp,imagereward}. 
However, the application of the reward model in score distillation of LiDAR scene completion faces following challenges. First, due to the complexity of LiDAR scenes, obtaining large-scale human-labeled data is challenging. With limited data, the reward model is easily over-optimized and faces the issue of reward hacking~\cite{hacking}. Second, existing methods often use differentiable rewards to optimize the model~\cite{clarkdirectly}, but commonly evaluation metrics such as IoU~\cite{IoU} and EMD~\cite{EMD} are non-differentiable and computationally expensive, difficult to be used directly as rewards to optimize the diffusion model. 

Compared to reward models, Diffusion-DPO~\cite{diffusiondpo,DPO} directly optimizes the diffusion model using preference data pairs, eliminating the need for training an additional reward model and thus avoiding the issue of reward hacking. 
Thus, to tackle the above challenges, we incoporate score distillation with the post training of DPO and propose a novel distillation framework dubbed Distillation-DPO for LiDAR scene completion diffusion models. 
Distillation-DPO includes an effective distillation strategy on the preference completed scene pairs for the first time. Specifically, based on the completed scene generated by the student model, we use LiDAR scene evaluation metrics as preference to construct the win-lose preference pairs. Then, Distillation-DPO optimizes the student model by computing the score function on both the student and teacher models. 
Compared with state-of-the-art (SOTA) LiDAR scene completion models, Distillation-DPO achieves significantly accelerated sampling for LiDAR completion diffusion models while delivering higher-quality completion results, setting a new SOTA performance (\cref{fig:teaser}).

Our contributions are summarized as follows: (1) We propose Distillation-DPO, a novel distillation framework for LiDAR scene completion diffusion models, which is the first to perform distillation based on preference data pairs.
(2) Compared to the existing state-of-the-art (SOTA) LiDAR scene completion models, Distillation-DPO achieves breakthroughs in both completion quality and speed.





\section{Preliminary}
\label{sec:pre}
\subsection{LiDAR scene completion diffusion model}
The goal of the LiDAR scene completion diffusion model $\boldsymbol{\epsilon}_\theta$ is to predict noise based on the given LiDAR sparse scan $\mathcal{P}$, enabling a step-by-step denoising process from an initial noisy sample $\mathcal{G}_T$ to obtain a dense scene reconstruction $\mathcal{G}_0$. In the existing SOTA model LiDiff~\cite{LiDiff}, the sampling step is often set to $50$. 

Given a input sparse scan $\mathcal{P}=\{\boldsymbol{p}^1,\boldsymbol{p}^2,...,\boldsymbol{p}^N\}$ and the ground truth $\mathcal{G} = \{\boldsymbol{p}^1,\boldsymbol{p}^2,...,\boldsymbol{p}^M\}$ (N $\ll$ M), the noisy point cloud $\mathcal{G}_t = \{\boldsymbol{p}_t^1,\boldsymbol{p}_t^2,...,\boldsymbol{p}_t^M\}$ can be calculated in a point-wise fashion~\cite{LiDiff}
\begin{equation}
    \boldsymbol{p}^{m}_{t} =\boldsymbol{p}^{m}+\left(\sqrt{\bar{\alpha}_{t}} \mathbf{0}+\sqrt{1-\bar{\alpha}_{t}} \boldsymbol{\epsilon_t}\right) =\boldsymbol{p}^{m}+ \sqrt{1-\bar{\alpha}_{t}} \boldsymbol{\epsilon_t}
    \label{eq:lidar_point_noisy}
\end{equation}
Here $p^m \in \mathbb{R}^3$ is the point cloud. Such a diffusion method is adopted because LiDAR data is large in scale, and directly applying traditional noise injection methods like DDPM~\cite{DDPM} would compress the LiDAR point cloud into a smaller range, leading to loss of details.

Due to the local diffusion method in~\cref{eq:lidar_point_noisy}, $\mathcal{G}_T$ can not be directly approximated by the Gaussian distribution. Given a sparse LiDAR scan $\mathcal{P}$, the point in $\mathcal{P}$ is first replicated $K$ times to obtain a dense scan $\mathcal{P}^* = \{\boldsymbol{p}^{1*},\boldsymbol{p}^{2*}, \ldots, \boldsymbol{p}^{M*}\}$. Then, the initial noisy point cloud $\mathcal{G}^*_T = \{\boldsymbol{p}^{1*}_T,\boldsymbol{p}^{2*}_T, \ldots, \boldsymbol{p}^{M*}_T\}$ is calculated by sampling a Gaussian noise for each $\boldsymbol{p}^{m*} \in \mathcal{P}^*$ based on~\cref{eq:lidar_point_noisy}. Finally, a step-by-step denoising process in~\cref{eq:reverse_process} is conducted to generate the completed scene $\mathcal{G}_0$. 
\begin{equation}
\label{eq:reverse_process}
    \mathcal{G}^{t-1}=\frac{1}{\sqrt{\alpha^{t}}}\left(\mathcal{G}^{t}-\frac{1-\alpha_{t}}{\sqrt{1-\bar{\alpha}^{t}}} \boldsymbol{\epsilon}_{\theta}\left(\mathcal{G}^{t}, \mathcal{P}, t\right)\right)+\sigma^{t} \boldsymbol{z}
\end{equation}

\subsection{Score distillation}
Score distillation shares the same motivation as this paper, aiming to make the few-step distribution of the student model as close as possible to the multi-step distribution of the teacher model. Let $p_{\eta}$ and $p_\theta$ be the distribution of the student model and the teacher model, separately. Score Distillaiton amis to minimize the following KL divergence
\begin{equation}
    \min _{\eta} D_{K L}\left(p_{\eta}\left(\boldsymbol{x}_0\right) \| p_\theta\left(\boldsymbol{x}_0 \right)\right)
\label{eq:SD_KL_divergence}
\end{equation}
Directly solving the optimization problem in~\cref{eq:SD_KL_divergence} is difficult. Thus, according to Theorem 1 in~\cite{prolificdreamer}, it is equivalent to the optimization problems in different timesteps $t$
\begin{equation}
    \min _{\eta} D_{K L}\left(p_{\eta,t}\left(\boldsymbol{x}_t\right) \| p_{\theta,t}\left(\boldsymbol{x}_t\right)\right)
\label{eq:SD_KL_divergence_t}
\end{equation}
Thus, the gradient of the student model can be written as
\begin{equation}
\label{eq:SD_gradient}
\begin{aligned}
    & \nabla_\eta D_{K L}\left(p_{\eta,t}\left(\boldsymbol{x}_t\right) \| p_{\theta,t}\left(\boldsymbol{x}_t\right)\right) \\ & = 
    \mathbb{E}_{t,\epsilon} \left[ \nabla_{\boldsymbol{x}_t}\log p_{\eta,t}\left(\boldsymbol{x}_t\right) - \nabla_{\boldsymbol{x}_t} \log p_{\theta,t}\left(\boldsymbol{x}_t\right)\right] \frac{\partial \boldsymbol{x}_t}{\partial \eta}
\end{aligned}
\end{equation}

Then, the score $\nabla_{\boldsymbol{x}_t} \log p_{\theta,t}\left(\boldsymbol{x}_t\right)$ can be approximated by the pre-trained diffusion model $\boldsymbol{\epsilon}_\theta$, and the score $\nabla_{\boldsymbol{x}_t}\log p_{\eta,t}\left(\boldsymbol{x}_t\right)$ can be approximated by an teaching assistant model $\boldsymbol{\epsilon}_\phi$ which trained on the generative samples of the student model with standard diffusion loss. Thus, the gradient in~\cref{eq:SD_gradient} can be approximated by 
\begin{equation}
\label{eq:SD_gradient_eps}
\begin{aligned}
    & \nabla_\eta D_{K L}\left(p_{\eta,t}\left(\boldsymbol{x}_t\right) \| p_{\theta,t}\left(\boldsymbol{x}_t\right)\right) \\ & \approx 
    \mathbb{E}_{t,\epsilon} \left[ \boldsymbol{\epsilon}_\theta(\boldsymbol{x}_t,t) - \boldsymbol{\epsilon}_\phi(\boldsymbol{x}_t,t)\right] \frac{\partial \boldsymbol{x}_t}{\partial \eta}
\end{aligned}
\end{equation}

During the training, the student model and the teaching assistant model $\boldsymbol{\epsilon}_\phi$ are optimized alternately. 

\subsection{A brief introduction of Diffusion-DPO}
This part reviews the Direct Preference Optimization in diffusion models (Diffusion-DPO)~\cite{diffusiondpo}. Let $\mathcal{D}=\{(\boldsymbol{c}, \boldsymbol{x}_0^w,\boldsymbol{x}_0^l\}$ is a dataset, where each data sample consists of a prompt $\boldsymbol{c}$ and a pair of images $\boldsymbol{x}_0^w$ and $\boldsymbol{x}_0^l$ with human preference $\boldsymbol{x}_0^w \succ \boldsymbol{x}_0^l$. The image $\boldsymbol{x}_0^w$ and $\boldsymbol{x}_0^l$ are both sampled from a references distribution $p_{\mathrm{ref}}$. To obtain the reward on the whole diffusion path, $r(\boldsymbol{c},\boldsymbol{x}_0)$ is defined as:
\begin{equation}
    r\left(\boldsymbol{c}, \boldsymbol{x}_{0}\right)=\mathbb{E}_{p_{\eta}\left(\boldsymbol{x}_{1: T} \mid \boldsymbol{x}_{0}, \boldsymbol{c}\right)}\left[R\left(\boldsymbol{c}, \boldsymbol{x}_{0: T}\right)\right]
    \label{eq:dpo_reward}
\end{equation}

Here $p_\eta$ is a diffusion model trained to align with human preferences. Then, $p_\eta$ can be optimized by maximizing the following objective
\begin{equation}
\begin{aligned}
    \begin{array}{l} \max _{p_{\eta}} \mathbb{E}_{\boldsymbol{c} \sim \mathcal{D}_{c}, \boldsymbol{x}_{0: T} \sim p_{\eta}\left(\boldsymbol{x}_{0: T} \mid \boldsymbol{c}\right)}\left[r\left(\boldsymbol{c}, \boldsymbol{x}_{0}\right)\right] \\ \quad-\beta \mathbb{D}_{\mathrm{KL}}\left[p_{\eta}\left(\boldsymbol{x}_{0 ; T} \mid \boldsymbol{c}\right) \| p_{\mathrm{ref}}\left(\boldsymbol{x}_{0: T} \mid \boldsymbol{c}\right)\right]\end{array}
\end{aligned}
\label{eq:dpo_objective}
\end{equation}

Compared to traditional DPO~\cite{DPO}, the objective function in~\cref{eq:dpo_objective} is defined over the entire diffusion path $\boldsymbol{x}_{0:T}$, which amis to maximize the reward $r\left(\boldsymbol{c}, \boldsymbol{x}_{0}\right)$ while ensuring that the distributions of $p_\eta$ and $p_{\mathrm{ref}}$ remain as close as possible. The objective in~\cref{eq:dpo_objective} can be further transformed into the following objective:
\begin{equation}
\begin{aligned}
    \begin{array}{l}L_{\mathrm{DPO-Diffusion}}(\eta)=-\mathbb{E}_{\left(\boldsymbol{x}_{0}^{w}, \boldsymbol{x}_{0}^{t}\right) \sim \mathcal{D}} \log \sigma \\ \left(\beta \mathbb{E}_{\substack{\boldsymbol{x}_{1: T}^{w} \sim p_{\eta}\left(\boldsymbol{x}_{1: T}^{w} \mid \boldsymbol{x}_{0}^{w}\right) \\ \boldsymbol{x}_{1: T}^{L} \sim p_{\eta}\left(\boldsymbol{x}_{1: T}^{l} \mid \boldsymbol{x}_{0}^{l}\right)}}\left[\log \frac{p_{\eta}\left(\boldsymbol{x}_{0: T}^{w}\right)}{p_{\text {ref }}\left(\boldsymbol{x}_{0: T}^{\omega}\right)}-\log \frac{p_{\eta}\left(\boldsymbol{x}_{0: T}^{l}\right)}{p_{\text {ref }}\left(\boldsymbol{x}_{0: T}^{l}\right)}\right]\right)\end{array}
\end{aligned}
\label{eq:dpo_theta}
\end{equation}

Here prompt $\boldsymbol{c}$ is omitted for compactness. 
By approximating the reverse process $p_{\eta}(\boldsymbol{x}_{1:T} | \boldsymbol{x}_0)$ with the forward process  $q(\boldsymbol{x}_{1:T} | \boldsymbol{x}_0)$, with some simplification, we have:


\begin{equation}
\begin{aligned}
L(\eta)= & -\mathbb{E}_{(\boldsymbol{x}_{0}^{w}, \boldsymbol{x}_{0}^{l}), t, \boldsymbol{x}_{t}^{w}, \boldsymbol{x}_{t}^{l}} \log \sigma(-\beta T \omega(\lambda_{t}) \\ &(\|\boldsymbol{\epsilon}^{w}-\boldsymbol{\epsilon}_{\eta}(\boldsymbol{x}_{t}^{w}, t)\|_{2}^{2}-\|\boldsymbol{\epsilon}^{w}-\boldsymbol{\epsilon}_{\mathrm{ref}}(\boldsymbol{x}_{t}^{w}, t)\|_{2}^{2} \\ & -(\|\boldsymbol{\epsilon}^{l}-\boldsymbol{\epsilon}_{\eta}(\boldsymbol{x}_{t}^{l}, t)\|_{2}^{2}-\|\boldsymbol{\epsilon}^{l}-\boldsymbol{\epsilon}_{\mathrm{ref}}(\boldsymbol{x}_{t}^{l}, t)\|_{2}^{2})))
\end{aligned}
\label{eq:dpo_loss}
\end{equation}

Here $\boldsymbol{x}_{t}^{*} = \alpha_t\boldsymbol{x}_{0}^{*} + \sigma_t \epsilon^*$, $\lambda_t = \frac{\alpha^2_t}{\sigma_t^2}$ is the signal-noise ratio, $\omega(\lambda_t)$ is the weighted function.
\section{Method}
\label{sec:method}

\begin{figure*}[t]
    \centering
    \includegraphics[width=\linewidth]{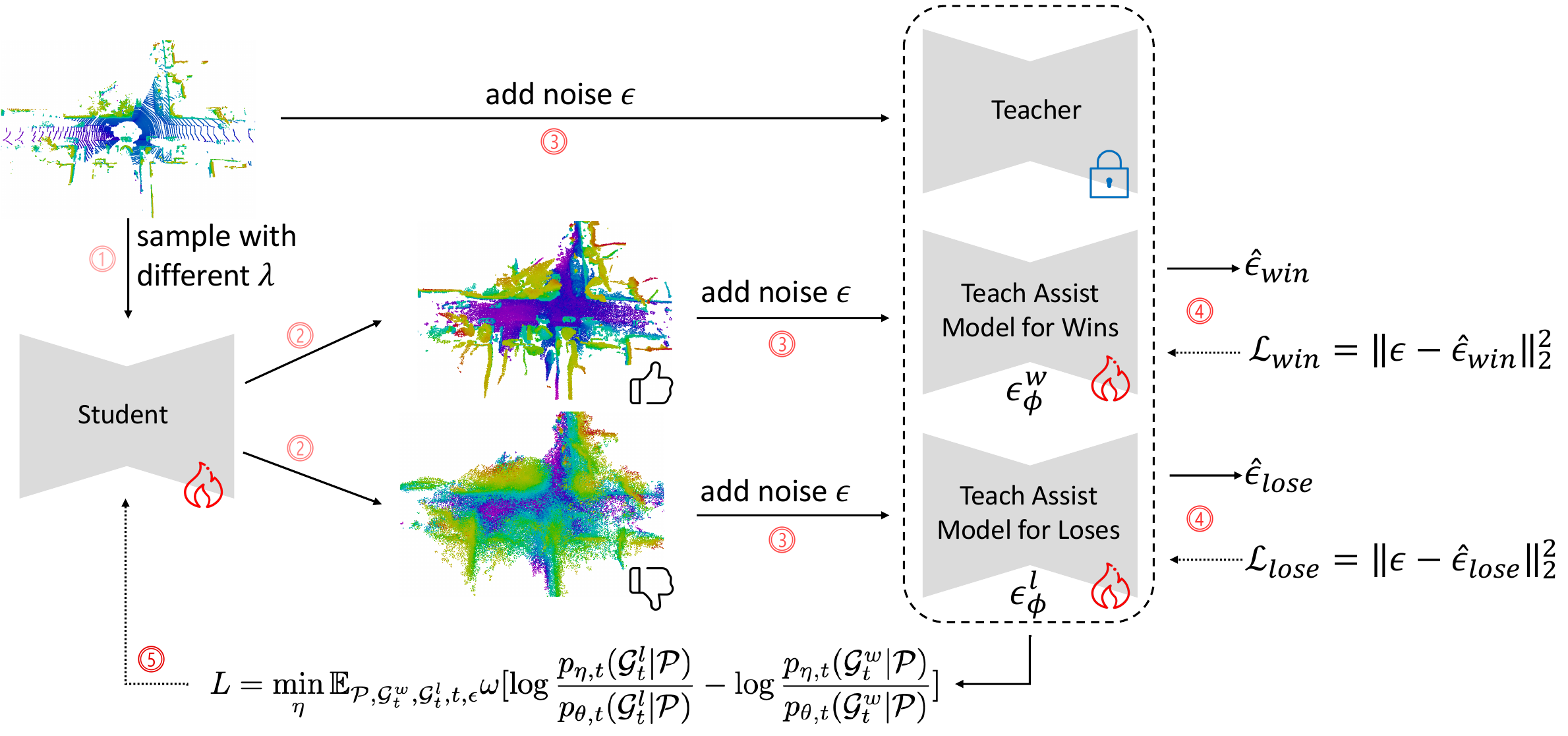}
    \caption{The overall structure of Distillation-DPO. (1) The student model generates the completed scene with different initial noise level $\lambda$ based on the sparse scan. (2) Choosing the winning sample $\mathcal{G}_t^w$ and losing samples $\mathcal{G}_t^l$. (3) The sparse scan, $\mathcal{G}_t^w$ and $\mathcal{G}_t^l$ are input to $\boldsymbol{\epsilon}_\theta$, $\boldsymbol{\epsilon}_\phi^w$ and $\boldsymbol{\epsilon}_\phi^l$.
    (4) The model $\boldsymbol{\epsilon}_\theta^w$ and $\boldsymbol{\epsilon}_\theta^l$ are optimized on $\mathcal{G}_t^w$ and $\mathcal{G}_t^l$, separately. (5) The student model is optimized by the DPO gradient.}
    \vspace{-1em}
    \label{fig:structure}
\end{figure*}

In this section, we introduce the proposed Distillation-DPO. Distillation-DPO aims to use preference-labeled data pairs to distill a pre-trained teacher LiDAR scene completion diffusion model into a student model, enabling the student model to achieve better completion results with fewer sampling steps. The overall structure of Distillation-DPO is shown in~\cref{fig:structure}.

As shown in~\cref{eq:dpo_objective}, Diffusion-DPO minimizes the KL divergence between the generative distribution $p_\theta$ and the reference distribution $p_\mathrm{ref}$ over the entire diffusion path $\boldsymbol{x}_{0:T}$. Therefore, the more sampling steps there are, the lower the efficiency of the optimization. Given spase scan $\mathcal{P}=\{\boldsymbol{p}^1,\boldsymbol{p}^2,...,\boldsymbol{p}^N\}$ and the completed scene $\mathcal{G}_0 = \{\boldsymbol{p}_0^1,\boldsymbol{p}_0^2,...,\boldsymbol{p}_0^M\}$, we first rewrite the optimization objective in~\cref{eq:dpo_objective} as:
\begin{equation}
    \min_\eta \mathbb{E}_{\mathcal{P}, \mathcal{G}_0}[D_{KL}(p_{\eta}(\mathcal{G}_0|\mathcal{P})) || p_\theta(\mathcal{G}_0|\mathcal{P})) - \omega r(\mathcal{G}_0,\mathcal{P})]
    \label{eq:scoredpo_objective}
\end{equation}

Here $p_\theta$ is the pre-trained distribution of the teacher model parameterized by $\theta$, $p_\eta$ is the generative distribution of $G_{stu}$ parameterized by $\eta$. The completed LiDAR scene $\mathcal{G}_0$ is generated by $G_{stu}$ with fewer inference steps based on the sparse LiDAR scan $\mathcal{P}$. However, directly optimizing~\cref{eq:scoredpo_objective} is challenging, because the high-density regions of $p_{\eta}(\mathcal{G}_0 | \mathcal{P})$ are sparse in high-dimensional spaces~\cite{prolificdreamer}. According to Theorem 1 in~\cite{prolificdreamer}, we extend~\cref{eq:scoredpo_objective} into an optimization over different time steps,
\begin{equation}
    \min_\eta \mathbb{E}_{\mathcal{P}, \mathcal{G}_t, \epsilon}[D_{KL}(p_{\eta,t}(\mathcal{G}_t|\mathcal{P})) || p_{\theta,t}(\mathcal{G}_t|\mathcal{P}))-\omega r(\mathcal{G}_t,\mathcal{P})]
    \label{eq:scoredpo_objective_t}
\end{equation}

Here $\epsilon$ is a random noise, $p_{\eta,t}$ and $p_{\theta,t}$ are the noisy distribution of the student model and the pre-trained teacher model at timestep $t$, separately. $\omega$ is the weight to control preference learning. Noisy completed LiDAR scene $\mathcal{G}_t=\{\boldsymbol{p}_t^1,\boldsymbol{p}_t^2,...,\boldsymbol{p}_t^M\}$ is obtained using the point-level noise addition method in~\cref{eq:lidar_point_noisy}. Using~\cref{eq:scoredpo_objective_t} and some algebra, the optimization problem can be written as
\begin{equation}
    \min_\eta \mathbb{E}_{\mathcal{P}, \mathcal{G}_0, \epsilon, t} [\log \frac{p_{\eta,t}(\mathcal{G}_t|\mathcal{P}) }{p_{\theta,t}(\mathcal{G}_t|\mathcal{P})} - \omega r(\mathcal{G}_t,\mathcal{P})]
    \label{eq:scoredpo_objective_E}
\end{equation}

For~\cref{eq:scoredpo_objective_E}, the global optimal solution $p_{\eta}^*$ is 
\begin{equation}
\begin{aligned}
     p_{\eta,t}^*(\mathcal{G}_t|\mathcal{P}) &= \frac{p_{\theta,t}(\mathcal{G}_t|\mathcal{P}) \exp(\omega r(\mathcal{G}_t,\mathcal{P}))}{Z(\mathcal{P})} \\
    Z(\mathcal{P}) &= \mathbb{E}_{\mathcal{G}_0,t,\epsilon} p_{\theta,t}(\mathcal{G}_t|\mathcal{P}) \exp(\omega r(\mathcal{G}_t,\mathcal{P})) 
\end{aligned}
\label{eq:global_optimal}
\end{equation}

Then, the reward function takes the form:
\begin{equation}
    r(\mathcal{G}_0,\mathcal{P}) = \frac{1}{\omega} \log\frac{p_{\eta,t}(\mathcal{G}_t|\mathcal{P})}{p_{\theta,t}(\mathcal{G}_t|\mathcal{P})} + \frac{1}{\omega}  \log Z(\mathcal{P})
\end{equation}

Hence, the objective of Distillation-DPO is obtained
\begin{equation}
    \min_\eta \mathbb{E}_{\mathcal{P}, \mathcal{G}_t^w, \mathcal{G}_t^l,t,\epsilon} \frac{1}{\omega}  [\log\frac{p_{\eta,t}(\mathcal{G}_t^l|\mathcal{P})}{p_{\theta,t}(\mathcal{G}_t^l|\mathcal{P})}-\log\frac{p_{\eta,t}(\mathcal{G}_t^w|\mathcal{P})}{p_{\theta,t}(\mathcal{G}_t^w|\mathcal{P})}]
    \label{eq:scoredpo_loss}
\end{equation}

Similarly, $\mathcal{G}_t^w$ and $\mathcal{G}_t^l$ represent the completed scenes by student model $G_{stu}$ with completion quality $\mathcal{G}_t^w \succ \mathcal{G}_t^l$. The gradient of $G_{stu}$ can be calculate as
\begin{equation}
\begin{aligned}
    \text{Grad}&(\eta) = \mathbb{E}_{\mathcal{P}, \mathcal{G}_t^w, \mathcal{G}_t^l,t,\epsilon}\\
    &  \frac{1}{\omega}  [( \nabla_{\mathcal{G}_t^l}\log p_{\eta,t}(\mathcal{G}_t^l|\mathcal{P}) - \nabla_{\mathcal{G}_t^l}\log p_{\theta,t}(\mathcal{G}_t^l|\mathcal{P}))\frac{\partial \mathcal{G}_t^l}{\partial \eta}- \\
    & ( \nabla_{\mathcal{G}_t^w}\log p_{\eta,t}(\mathcal{G}_t^w|\mathcal{P}) - \nabla_{\mathcal{G}_t^w}\log p_{\theta,t}(\mathcal{G}_t^w|\mathcal{P}))\frac{\partial \mathcal{G}^w_t}{\partial \eta}]
\end{aligned}
\label{eq:scoredpo_gradient}
\end{equation}

Score $\nabla_{\mathcal{G}_t^l}\log p_{\theta,t}(\mathcal{G}_t^w|\mathcal{P})$ and $\nabla_{\mathcal{G}_t^l}\log p_{\theta,t}(\mathcal{G}_t^l|\mathcal{P})$ is approximated by the pre-trained teacher diffusion model $\boldsymbol{\epsilon}_\theta$. Differently, the score $\nabla_{\mathcal{G}_t^w}\log p_{\eta,t}(\mathcal{G}_t^w|\mathcal{P})$ and $\nabla_{\mathcal{G}_t^l}\log p_{\eta,t}(\mathcal{G}_t^w|\mathcal{P})$ is approximated by two teaching assistant models $\boldsymbol{\epsilon}_\phi^w$ and $\boldsymbol{\epsilon}_\phi^l$. Therefore, the gradient of $G_{stu}$ is
\begin{equation}
\begin{aligned}
    & \text{Grad}(\eta) \\
    = & \mathbb{E}_{\mathcal{P}, \mathcal{G}_t^w, \mathcal{G}_t^l,t,\epsilon} \frac{1}{\omega}  [(\boldsymbol{\epsilon}_\theta(\mathcal{G}_t^l,t,\mathcal{P}) - \boldsymbol{\epsilon}_\phi(\mathcal{G}_t^l,t,\mathcal{P}))\frac{\partial \mathcal{G}^l_t}{\partial \eta}\\
    - &(\boldsymbol{\epsilon}_\theta(\mathcal{G}_t^w,t,\mathcal{P}) - \boldsymbol{\epsilon}_\phi(\mathcal{G}_t^w,t,\mathcal{P}))\frac{\partial \mathcal{G}^w_t}{\partial \eta}]
\end{aligned}
\label{eq:scoredpo_gradient_dm}
\end{equation}

To generate preference-aware completed scenes $\mathcal{G}_0^w$ and $\mathcal{G}_0^l$, we first introduce a parameter $\lambda$ when computing $\mathcal{G}_T$, which controls the initial noise scale,
\begin{equation}
    \boldsymbol{p}^{m}_{T} =\boldsymbol{p}^{m}+\lambda\sqrt{1-\bar{\alpha}_{T}} \boldsymbol{\epsilon_T}
    \label{eq:lidar_point_noisy_T}
\end{equation}

By default, $\lambda = 1$. To generate completed scene $\mathcal{G}_0^w$ and $\mathcal{G}_0^l$ separately based on the same sparse scan $\mathcal{P}$, we obtain different completion results by adjusting different values of $\lambda$. We set $\lambda > 1$ to obtain a $\mathcal{G}_T^\prime$ different from $\mathcal{G}_T$, which is then used to generate $\mathcal{G}_0^\prime$ different from $\mathcal{G}_0$. Then, according to the completion quality metrics, we assign the sample with the higher quality as $\mathcal{G}_0^w$ and another as $\mathcal{G}_0^l$.


During the training process, the student model $G_{stu}$ and two teaching assistant models $\boldsymbol{\epsilon}_\phi^w$ and $\boldsymbol{\epsilon}_\phi^l$ are optimized alternately. The teaching assistant models $\boldsymbol{\epsilon}_\phi^w$ and $\boldsymbol{\epsilon}_\phi^l$ are trained on the completed scene generated by $G_{stu}$ with the standard diffusion objective~\cite{DDPM}  
\begin{equation}
 \mathcal{L}_{DM}=\mathbb{E}_{\mathcal{P},t, \epsilon} \left[\left\|\boldsymbol{\epsilon}-\boldsymbol{\epsilon}_{\phi}^i\left(\mathcal{G}_{t}^i, \mathcal{P}, t\right)\right\|^{2}\right] \quad i \in \{w,l\}
\label{eq:lidar_dm_loss}
\end{equation}
\section{Experiment}
\label{sec:exp}
\paragraph{Model and datasets}
We use the existing SOTA 3D LiDAR scene completion diffusion model LiDiff~\cite{LiDiff} as the teacher and train a few-step student model with~\cref{eq:scoredpo_gradient_dm}. LiDiff can achieve complete a scene with 50 sampling steps based on the sparse LiDAR scan. The student model $G$ and the teaching assistant models $\boldsymbol{\epsilon}_\phi^w$ and $\boldsymbol{\epsilon}_\phi^l$ are initialized with the pre-trained LiDiff model, but the student model performs scene completion with fewer sampling steps. The experiments are conducted on the SemanticKITTI~\cite{semantickitti} dataset.

\paragraph{Baselines and metrics}
Except for the existing SOTA LiDAR scene completion diffusion model LiDiff~\cite{LiDiff}, we also choose LMSCNet~\cite{lmscnet}, LODE~\cite{lode}, MID~\cite{MID} and PVD~\cite{pvd} as the baselines. We evaluate the performance of the proposed Distillation-DPO on Chamfer Distance (CD)~\cite{CD}, Jensen-Shannon Divergence (JSD)~\cite{JSD} and Earth Mover's Distance (EMD)~\cite{EMD}. These three metrics can provide a comprehensive evaluation of the completed LiDAR scene quality from different perspectives.

\subsection{Evaluation on LiDAR scene completion}
\label{sec:quantitative}
We first compared the performance of the proposed Distillation-DPO and existing models in LiDAR scene completion on the SemanticKITTI dataset. According to different settings, Distillation-DPO can perform sampling with different inference steps. As the sampling steps decrease, the scene completion speed increases, but it inevitably sacrifices some completion quality. After balancing completion speed and quality, we chose the result with 8 sampling steps as the completion output of Distillation-DPO for comparison with existing models. In~\cref{sec:ablation}, we further compare the performance of Distillation-DPO under different sampling steps.

The comparison results of Distillation-DPO are shown in~\cref{tab:completion1}. Distillation-DPO achieves the optimal completion quality except in EMD. Compared with the SOTA LiDAR scene completion method LiDiff~\cite{LiDiff}, Distillation-DPO accelerates the completion speed by over $5$ times while achieving improvements of $6\%$ and $7\%$ in CD and JSD. As for EMD, Distillation-DPO still maintains a comparable performance compared with the existing method. Although the sampling speed of Distillation-DPO is slower than LMSCNet~\cite{lmscnet}, LODE~\cite{lode}, the sampling quality has been significantly improved.

\begin{table}[t]
\centering
\resizebox{\columnwidth}{!}{%
\begin{tabular}{lrrrr}
\toprule
Model & \multicolumn{1}{c}{CD $\downarrow$} & \multicolumn{1}{c}{JSD $\downarrow$} & \multicolumn{1}{c}{EMD $\downarrow$} & \multicolumn{1}{c}{Times (s) $\downarrow$} \\ \midrule
LMSCNet$^{\dag}$~\cite{lmscnet} & 0.641 & 0.431 & - & - \\
LODE$^{\dag}$~\cite{lode} & 1.029 & 0.451 & - & - \\
MID$^{\dag}$~\cite{MID} & 0.503 & 0.470 & - & - \\
PVD~\cite{pvd} & 1.256 & 0.498 & - & - \\
LiDiff$^{\dag}$~\cite{LiDiff} & 0.434 & 0.444 & \cellcolor[RGB]{\colorfirst}22.15 & \cellcolor[RGB]{\colorthird}17.75 \\
LiDiff (Refined)$^{\dag}$~\cite{LiDiff} &  \cellcolor[RGB]{\colorsecond}0.375 &  \cellcolor[RGB]{\colorsecond}0.416 & \cellcolor[RGB]{\colorsecond}23.16 & 17.87 \\ 
\midrule
Distillation-DPO & \cellcolor[RGB]{\colorthird}0.414 & \cellcolor[RGB]{\colorthird}0.419 & \cellcolor[RGB]{\colorthird}23.29 & \cellcolor[RGB]{\colorfirst}3.28\\
Distillation-DPO (Refined) & \cellcolor[RGB]{\colorfirst}0.354 & \cellcolor[RGB]{\colorfirst}0.387 &  23.66 & \cellcolor[RGB]{\colorsecond}3.38 \\ \bottomrule
\end{tabular}%
}
\caption{The results on LiDAR scene completion of Distillation-DPO with existing models. Colors denote the \colorbox[RGB]{\colorfirst}{1st}, \colorbox[RGB]{\colorsecond}{2nd}, and \colorbox[RGB]{\colorthird}{3rd} best-performing model. ``$\dag$'' means the completion time is calculated based on the official implementation and released checkpoints. Here Lidiff takes 50 NFEs while ours takes 8 only.}
\label{tab:completion1}
\end{table}

\subsection{Ablation study}
\label{sec:ablation}
In this part, we first show the completion results of Distillation-DPO with different inference steps. \cref{tab:ablation1} shows the results. As the number of inference steps decreases, the completion speed of Distillation-DPO is further reduced. With just one sampling step, it only takes $0.69$ seconds to complete a scene. However, the reduction in inference steps leads to a decline in completion quality. The speed improvement gained from sampling steps reduction is not enough to compensate for the loss in quality. Therefore, choosing 8 steps by default is a good balance of speed and efficiency.

\begin{table}[t]
\centering
\resizebox{\columnwidth}{!}{%
\begin{tabular}{lrrrrr}
\toprule
Model & \multicolumn{1}{c}{NFE $\downarrow$} & \multicolumn{1}{c}{CD $\downarrow$} & \multicolumn{1}{c}{JSD $\downarrow$} & \multicolumn{1}{c}{EMD $\downarrow$} & \multicolumn{1}{c}{Time (s) $\downarrow$} \\ \midrule
LiDiff~\cite{LiDiff} & 50 & 0.434 & 0.444 & \cellcolor[RGB]{\colorfirst}22.15 & 17.75 \\
LiDiff (Refined)~\cite{LiDiff} & 50 & \cellcolor[RGB]{\colorsecond}0.375 & 0.416 & \cellcolor[RGB]{\colorsecond}23.16 & 17.87 \\
LiDiff~\cite{LiDiff} & 8 & 0.447 & 0.432 & 24.90 & 3.35 \\
LiDiff (Refined)~\cite{LiDiff} & 8 & \cellcolor[RGB]{\colorthird}0.411 & \cellcolor[RGB]{\colorthird}0.406 & 25.74 & 3.48 \\ \midrule
Distillation-DPO (Refined) & 8 & \cellcolor[RGB]{\colorfirst}0.354 & \cellcolor[RGB]{\colorfirst}0.387 & \cellcolor[RGB]{\colorthird}23.66 & 3.38 \\
Distillation-DPO (Refined) & 4 & 0.429 & 0.413 & 24.24 & \cellcolor[RGB]{\colorthird}1.84 \\
Distillation-DPO (Refined) & 2 & 0.475 & \cellcolor[RGB]{\colorsecond}0.398 & 25.30 & \cellcolor[RGB]{\colorsecond}1.08\\
Distillation-DPO (Refined) & 1 & 0.645 & 0.430 & 28.11 & \cellcolor[RGB]{\colorfirst}0.69 \\ \bottomrule
\end{tabular}%
}
\caption{Comparison results of different inference steps on the SemanticKITTI dataset.}
\label{tab:ablation1}
\end{table}

Then, we further compare the completion quality of different values of $\lambda$. In the implementation of Distillation-DPO, we set $\lambda=1.1$ by default to calculate $\mathcal{G}_T^\prime$. Here, we use different $\lambda$ values to train Distillation-DPO and compare the results in~\cref{tab:ablation2}. When decreasing or increasing $\lambda$, the completion performance deteriorates. When $\lambda$ is small, the difference between $\mathcal{G}_0^w$ and $\mathcal{G}_0^l$ is minimal, making the gradients of the student model in~\cref{eq:scoredpo_gradient_dm} small, which leads to unstable training. Conversely, when $\lambda$ is large, the quality of $\mathcal{G}_0^\prime$ generated from $\mathcal{G}_T^\prime$ degrades significantly, causing it to fall outside the distribution learned by the pre-trained teacher model $\boldsymbol{\epsilon}_{\theta}$. This mismatch leads to inaccurate predictions from $\boldsymbol{\epsilon}_{\theta}$~\cite{ScoreLidAR}, resulting in incorrect gradients for the student model and ultimately lowering the completion quality.
\begin{table}[t]
\centering
\begin{tabular}{lrrr}
\toprule
Model & \multicolumn{1}{c}{CD $\downarrow$} & \multicolumn{1}{c}{JSD $\downarrow$} & \multicolumn{1}{c}{EMD $\downarrow$}  \\ \midrule
$\lambda = 1.1$ (ours) & \cellcolor[RGB]{\colorfirst}0.354 & \cellcolor[RGB]{\colorfirst}0.387 & 23.66 \\
$\lambda = 1.05$ & \cellcolor[RGB]{\colorthird}0.418 & \cellcolor[RGB]{\colorsecond}0.421 & \cellcolor[RGB]{\colorsecond}23.48 \\
$\lambda = 1.2$ & 0.421 & 0.423 & \cellcolor[RGB]{\colorfirst}23.44 \\
$\lambda = 1.5$ & \cellcolor[RGB]{\colorsecond}0.409 & \cellcolor[RGB]{\colorthird}0.422 & \cellcolor[RGB]{\colorthird}23.60 \\ 
$\lambda = 2.0$ & 0.427 & 0.432 & 23.82 \\ \bottomrule
\end{tabular}%
\caption{Comparison results of different $\lambda$ value SemanticKITTI dataset. All results have been refined.}
\label{tab:ablation2}
\end{table}

We also conducted experiments to explore the impact of different teacher model performances on the effectiveness of Distillation-DPO. Theoretically, the final performance of the student model is constrained by the teacher model. The better the performance of the teacher model, the better the final performance of the student model. Thus, we first fine-tuned LiDiff~\cite{LiDiff} using  DiffusionDPO~\cite{diffusiondpo} to enhance its performance. Then, we retrained Distillation-DPO using the fine-tuned model. Results shown in~\cref{tab:ablation3} display that as the performance of the teacher model improves, the performance of Distillation-DPO also improves.
\begin{table}[t]
\centering
\begin{tabular}{lrrr}
\toprule
Model & \multicolumn{1}{c}{CD $\downarrow$} & \multicolumn{1}{c}{JSD $\downarrow$} & \multicolumn{1}{c}{EMD $\downarrow$}  \\ \midrule
LiDiff~\cite{LiDiff} & 0.375 & 0.416 & 23.16 \\
LiDiff$^*$ & 0.368 & 0.401 & \cellcolor[RGB]{\colorfirst}22.69 \\
Distillation-DPO & 0.354 & 0.387 & 23.66 \\
Distillation-DPO$^*$ & \cellcolor[RGB]{\colorfirst}0.343 & \cellcolor[RGB]{\colorfirst}0.385 & 23.53 \\ \bottomrule
\end{tabular}%
\caption{Comparison results of using different teacher models. LiDiff$^*$ represents the LiDiff model refined with Diffusion-DPO and it enjoys boosted performance. Distillation-DPO$^*$ represents the Distillation-DPO trained with LiDiff$^*$. With a stronger teacher, the distillated student also have better performance. All results have been refined.}
\label{tab:ablation3}
\end{table}

Moreover, we conduct experiments by changing the evaluation metric for determining $\mathcal{G}_0^w$ and $\mathcal{G}_0^l$ to JSD. The results in~\cref{tab:ablation4} show that the performance significantly deteriorates when using JSD. Since JSD measures the similarity of point cloud distributions, it requires a large number of samples to estimate the probability density distribution accurately. However, when comparing and determining whether a sample is $\mathcal{G}_0^w$ and $\mathcal{G}_0^l$, the metric is computed using only a single generated sample and its corresponding ground truth. In this case, JSD becomes inaccurate and may even lose its practical significance, leading to the performance decline.
\begin{table}[t]
\centering
\begin{tabular}{lrrr}
\toprule
Model & \multicolumn{1}{c}{CD $\downarrow$} & \multicolumn{1}{c}{JSD $\downarrow$} & \multicolumn{1}{c}{EMD $\downarrow$}  \\ \midrule
Distillation-DPO (CD) & \cellcolor[RGB]{\colorfirst}0.354 & \cellcolor[RGB]{\colorfirst}0.387 & \cellcolor[RGB]{\colorfirst}23.66 \\
Distillation-DPO (JSD) & 0.444 & 0.445 & 24.82 \\ \bottomrule
\end{tabular}%
\caption{Comparison results of using different metrics to determine $\mathcal{G}_0^w$ and $\mathcal{G}_0^l$. All results have been refined.}
\label{tab:ablation4}
\end{table}

Finally, we further compare Distillation-DPO with results distilled using traditional score distillation methods to validate the effectiveness of the proposed distillation framework. \cref{tab:ablation5} shows that the results obtained using score distillation are even inferior to those of the original teacher model LiDiff~\cite{LiDiff}. This is consistent with our statement in~\cref{sec:intro} that directly employing score distillation can accelerate the sampling speed while inevitably leading to a drop in performance. In contrast, the proposed Distillation-DPO distillation framework incorporates guidance from preference data, which not only accelerates sampling but also further enhances completion quality, thereby achieving efficient and high-quality scene completion.
\begin{table}[t]
\centering
\begin{tabular}{lrrr}
\toprule
Model & \multicolumn{1}{c}{CD $\downarrow$} & \multicolumn{1}{c}{JSD $\downarrow$} & \multicolumn{1}{c}{EMD $\downarrow$}  \\ \midrule
LiDiff~\cite{LiDiff} & \cellcolor[RGB]{\colorsecond}0.375 & \cellcolor[RGB]{\colorsecond}0.416 & \cellcolor[RGB]{\colorfirst}23.16 \\
Score Distillation & \cellcolor[RGB]{\colorthird}0.419 & \cellcolor[RGB]{\colorthird}0.430 & \cellcolor[RGB]{\colorthird}24.61 \\
Distillation-DPO & \cellcolor[RGB]{\colorfirst}0.354 & \cellcolor[RGB]{\colorfirst}0.387 & \cellcolor[RGB]{\colorsecond}23.66 \\ \bottomrule
\end{tabular}%
\caption{Comparison between Distillation-DPO and traditional score distillation. All results have been refined.}
\label{tab:ablation5}
\end{table}

\subsection{Qualitative comparison}
\begin{figure*}[t]
    \centering
    \includegraphics[width=\linewidth]{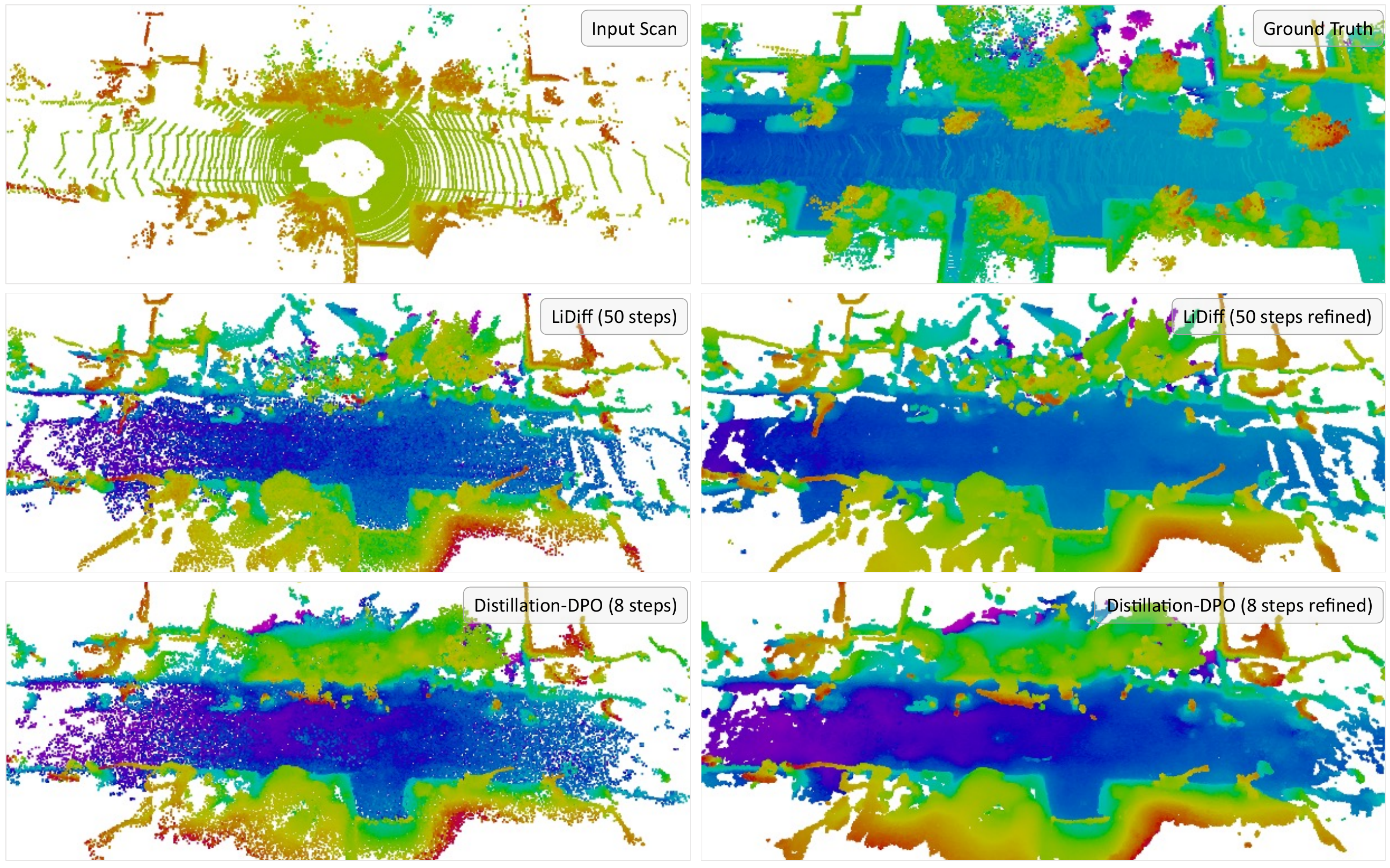}
    \caption{Qualitative results on SemanticKITTI. Compared to LiDiff~\cite{LiDiff}, Ditillation-DPO achieves faster and higher-quality completion.}
    \vspace{-1em}
    \label{fig:qualitative}
\end{figure*}

We visualized the scene completion results of Distillation-DPO and compared them with those of the SOTA model LiDiff~\cite{LiDiff}, as shown in the~\cref{fig:qualitative}. Compared to LiDiff, Distillation-DPO achieves higher scene completion quality with only 8 sampling steps, surpassing LiDiff’s results even with 50 sampling steps. Moreover, Distillation-DPO provides more complete reconstructions of fine details, such as cars, road cones, and signposts.

\label{sec:qualitative comparison}

\section{Discussion}
\label{sec:dis}
\subsection{Rationality of the student model initialization}
As in~\cref{sec:exp}, the student model $G$ is initialized from the pre-trained teacher diffusion model $\epsilon_\theta$.  This initialization approach is feasible and commonly used in existing methods~\cite {DMD,DMD2,Diff-Instruct}. Although the student model and the teacher model share the same initial parameters, the student model uses fewer sampling steps than the teacher model. As a result, at the beginning, the student model performs worse in few-step sampling compared to the teacher model’s multi-step sampling. The generated distribution of the student model differs from the pre-trained distribution of the teacher model. The objective of Distillation-DPO aligns with that of traditional score distillation: to ensure that the few-step sampling distribution of the student model closely matches the multi-step sampling distribution of the teacher model. This allows the student model to achieve comparable or even high performance with fewer sampling steps.

\subsection{Similarities and differences with Diffusion-DPO}
\paragraph{Similarities}
Both Diffusion-DPO~\cite{diffusiondpo} and the proposed Distillation-DPO use preference data pairs to optimize the model and maximize the reward.


\paragraph{Differences} First, the minimized KL divergences are different.
\begin{itemize}
    \item \textbf{Diffusion-DPO} minimizes the joint distribution over the entire diffusion path $\boldsymbol{x}_{0:T}$, i.e., the KL divergence between generative distribution $p_\eta(\boldsymbol{x}_{0:T}|c)$  and the reference distribution $p_{\mathrm{ref}}(\boldsymbol{x}_{0:T}|c)$.
    \item \textbf{Distillation-DPO} minimizes the KL divergence between the student model's generative distribution and the training distribution, which is reformulated as the KL divergence between the noised distributions at different timestep $t$. Since the training distribution is not accessible, Distillation-DPO approximates the training distribution using a pre-trained diffusion model. Therefore, the optimization objective of Distillation-DPO is transformed into minimizing the KL divergence between the student model’s generative distribution $p_\eta(\mathcal{G}_t|\mathcal{P})$ and the teacher model’s pre-trained distribution $p_\theta(\mathcal{G}_t|\mathcal{P})$.
\end{itemize}
Second, the optimization strategies are different.
\begin{itemize}
    \item \textbf{Diffusion-DPO} directly optimizes the generative model $p_\eta$ by~\cref{eq:dpo_loss}.
    \item \textbf{Distillation-DPO} first calculates the score difference of the winning sample $\mathcal{G}^w_0$ between the teaching assistant model $\boldsymbol{\epsilon}_\phi^w$ and the teacher model $\boldsymbol{\epsilon}_\theta$, as well as the score difference of the losing sample $\mathcal{G}^l_0$ between the teaching assistant model $\boldsymbol{\epsilon}_\phi^l$ and the teacher model $\boldsymbol{\epsilon}_\theta$. These two components are then combined as the gradient to optimize the student model $G_{stu}$. The teaching assistant models $\boldsymbol{\epsilon}_\phi^w$ and $\boldsymbol{\epsilon}_\phi^l$ are optimized separately on $\mathcal{G}^w_0$ and $\mathcal{G}^w_l$ based on the diffusion loss. The student $G_{stu}$ and the teaching assistant models are optimized alternately.
\end{itemize}
Third, the training policies are different.
\begin{itemize}
    \item \textbf{Diffusion-DPO}'s training is off-policy. Diffusion-DPO samples the preference data pair from a reference distribution $p_{\mathrm{ref}}$, which are predefined before training begins and remain unchanged throughout the training process.
    \item \textbf{Distillation-DPO}'s training is on-policy. Distillation-DPO generates the preference data pairs by the student model in each optimization step, which is changed with the optimization of the student model during the training.
\end{itemize}
Finally, the sampling steps are different.
\begin{itemize}
    \item \textbf{Diffusion-DPO} requires the sampling steps of the generative model to be consistent with those of the reference model during the training.
    \item \textbf{Distillation-DPO} does not require the student model to have the same number of sampling steps as the teacher model during training. The student model directly conducts single-step sampling during the training. 
\end{itemize}

\subsection{Similarities and differences with Score Distillation}
\paragraph{Similarities}
Both methods share the training objective of making the few-step distribution of the student model as close as possible to the multi-step distribution of the teacher.

\paragraph{Differences} The training strategies are different.
\begin{itemize}
    \item \textbf{Score Distillation} training the student model by the difference between two score functions without preference data pairs. Score Distillation only has one teaching assistant model $\boldsymbol{\epsilon}_\phi$ to approximate the generative distribution of the student model.
    \item \textbf{Distillation-DPO} calculates the score function differences on preference data pairs $\mathcal{G}_0^w$ and $\mathcal{G}_0^l$ separately and combines two terms as the gradient to optimize the student model. Distillation-DPO has two teaching assistant model $\boldsymbol{\epsilon}_\phi^w$ and $\boldsymbol{\epsilon}_\phi^l$ to approximate the distribution of $\mathcal{G}_0^w$ and $\mathcal{G}_0^l$ separately.
\end{itemize}

\subsection{Differential Rewards}

The rationale for employing 3D LiDAR scene completion evaluation metrics as preference signals stems from three compelling arguments.  
{First}, our approach diverges from direct reward optimization frameworks by leveraging these metrics exclusively for constructing preference-based {winning-losing pairs}. This methodology emulates human preference annotation paradigm, where evaluative criteria guide pairwise comparisons rather than serving as differentiable objectives. Such indirect alignment circumvents the risk of metric exploitation.
{Second}, prior works have demonstrated the feasibility of post-training optimization or test-time metric maximization using these criteria~\cite{clarkdirectly,singhal2025general,ma2025inference}, with reported performance gains validating their efficacy. Our methodology extends this convention through indirect utilization.  
{Third}, experiments employing Chamfer Distance and Jensen-Shannon Divergence as training signals demonstrated consistent performance improvements across other evaluation metrics. 

\section{Related Work}
\label{sec:rel}
\subsection{Preference Optimization for Diffusion Models}
To generate results that better align with human preferences, some studies have begun to train models based on preference-optimization methods~\cite{feedback,rich,DPO}. ImageReward~\cite{imagereward} proposes the first general human preference reward model for text-to-image generation and directly optimizes the diffusion model based on feedback during random subsequent denoising steps. Subsequent studies have further leveraged more detailed annotation methods~\cite{rich} and combined multiple open-source models~\cite{itercomp} to obtain richer human feedback datasets. Additionally, some works have optimized reward feedback learning by integrating multiple reward models~\cite{versat2i} or improving training methodologies~\cite{itercomp}. Since obtaining large-scale human annotations is challenging, some methods have attempted to train reward models using semi-supervised learning with unlabeled data~\cite{semi} or employing hybrid annotation strategies with AI and human~\cite{GRM}. Additionally, Diffusion-DPO~\cite{diffusiondpo} is the first to extend Direct Preference Optimization~\cite{DPO} to diffusion models, directly optimizing the model based on image preferences to eliminate the complex reward modeling and improve training efficiency.

\subsection{LiDAR Scene Completion}
LiDAR scene completion aims to reconstruct sparse LiDAR scans into dense and complete 3D point cloud scenes~\cite{ScoreLidAR}. Traditional LiDAR scene completion methods recover dense depth maps from sparse point clouds~\cite{depth1,depth2}, leveraging guidance from RGB images or bird’s-eye view images to achieve high-quality completion~\cite{chen2019learning,depth8}. Some methods represent LiDAR scenes as voxels and utilize Signed Distance Fields (SDFs) to reconstruct complete point cloud scenes~\cite{lode}. However, the completion quality of these methods is constrained by the voxel resolution~\cite{LiDiff}. Due to the high generative quality and strong training stability, many studies have recently leveraged diffusion models for high-quality LiDAR scene completion~\cite{diffusionlidar,diffssc,R2DM,ScoreLidAR}. Some methods focus on reconstructing sparse LiDAR scans into dense scans, such as R2DM~\cite{R2DM}, OLiDM~\cite{olidm}, and LiDMs~\cite{LiDMs}. Other approaches attempt to directly recover complete point cloud scenes from sparse LiDAR scans, including LiDiff~\cite{LiDiff} and DiffSSC~\cite{diffssc}. To further accelerate LiDAR scene completion speed, ScoreLiDAR introduces a distillation method based on structural loss, enabling fast and efficient LiDAR point cloud completion~\cite{ScoreLidAR}.
\section{Conclusion}
\label{sec:con}
\paragraph{Summary} This paper proposes a novel LiDAR scene completion diffusion model distillation framework, Distillation-DPO. Distillation-DPO redefines the Diffusion-DPO framework by introducing the score distillation strategy, enabling effective distillation of LiDAR scene completion diffusion models using preference data pairs. Compared to existing models, Distillation-DPO achieves new SOTA completion performance while improving completion speed more than five times over existing SOTA models.
To our best knowledge, we are the first to integrate distillation and post-training with preference and provide insight to preference-aligned diffusion distillation for both areas of LiDAR scene completion and visual generation. 

\paragraph{Limitation} Since the official implementations and models of SOTA diffusion-based semantic scene completion (SSC) models, such as DiffSSC~\cite{diffssc}, are not yet publicly available, Distillation-DPO has not yet been evaluated on the SSC task. Future work will explore its application in the SSC task. Additionally, while Distillation-DPO improves the sampling speed of existing models by over 5 times, it still does not achieve real-time LiDAR scene completion. Future work will focus on further accelerating the completion process without compromising quality, aiming to achieve real-time high-quality scene completion.
{
    \small
    \bibliographystyle{ieeenat_fullname}
    \bibliography{main}
}

\end{document}


\maketitle
\tableofcontents
\section{Experiment details}
\subsection{Dataset setup}
SemanticKITTI~\cite{semantickitti} is a sub-dataset of KITTI specifically designed for semantic segmentation and serves as a key benchmark for LiDAR-based semantic segmentation. It is one of the largest datasets with sequential information. SemanticKITTI provides dense per-point annotations for all sequences in the KITTI Vision Odometry Benchmark, covering the full 360$\degree$ field of view of the onboard LiDAR sensor. The dataset includes 28 labeled categories, divided into static and dynamic objects, encompassing pedestrians, vehicles, and other traffic participants, as well as ground infrastructure, such as parking lots and sidewalks. SemanticKITTI consists of 22 sequences with over 43,000 LiDAR scans and is commonly used for (i) Semantic segmentation of point clouds from single scans, (ii) Sequence-based segmentation, leveraging multiple past scans, and (iii) Semantic scene completion tasks.

\subsection{Implentation details}
All models in the experiments are based on the same network architecture as that of LiDiff~\cite{LiDiff}. The teacher model $\epsilon_{ref}$, teaching assistant models, $\epsilon_\phi^w$, $\epsilon_\phi^l$, and the student model $G_{stu}$, are all initialized using the pre-trained weights from LiDiff. The Distillation-DPO model is trained and evaluated on the SemanticKITTI dataset. Sequences $00$ to $07$ and $09$ to $10$ are used for training, while sequence $08$ serves as the validation set. During training, we use the Chamfer Distance as the metrics to decide the winning sample $\mathcal{G}^w_0$ and the losing sample $\mathcal{G}^l_0$ and use the Jensen-Shannon Divergence to conduct the ablation study. We employ the Stochastic Gradient Descent (SGD) optimizer with default parameters. The initial learning rate is set to $1e-5$, and we apply an Exponential Decay Scheduler with $\gamma = 0.999$, which results in the learning rate being annealed at each training iteration. At each iteration, the student model and the teaching assistant models are trained alternately, with each model updated exactly once per iteration. The initial noise level $\lambda$, which is an empirical parameter, is set to $1.1$ throughout the training process. All experiments are conducted using two Nvidia A100 GPUs and the PyTorch Lightning framework.

\section{Theoretical demonstration}
In this part, we provide the derivation of the global optimal solution $p_\eta^*$ in Eq.~\textcolor{red}{(15)} in the main paper. Recall that the optimization objective of Distillation-DPO is 
\begin{equation}
    \min_\eta \mathbb{E}_{\mathcal{P}, \boldsymbol{x}_t, \epsilon}[D_{KL}(p_{\eta,t}(\mathcal{G}_t|\mathcal{P})) || p_{\theta,t}(\mathcal{G}_t|\mathcal{P}))-\omega r(\mathcal{G}_t,\mathcal{P})]
    \label{eq:SM_scoredpo_objective_t}
\end{equation}

Given any reward model $(\mathcal{G}_t,\mathcal{P})$, teacher model $p_\theta$ and a general non-parametric policy class, we can obtain
\begin{equation}
\begin{aligned} 
    & \min _{\eta} \mathbb{E}_{\mathcal{P}, \mathcal{G}_t, \epsilon} {\beta \mathbb{D}_{\mathrm{KL}}\left[p_\eta(\mathcal{G}_t \mid \mathcal{P}) \| p_\theta(\mathcal{G}_t \mid \mathcal{P})\right] - \omega r(\mathcal{G}_t, \mathcal{P})} \\ & =\min _{\eta} \mathbb{E}_{\mathcal{P}} \mathbb{E}_{\mathcal{G}_t,\epsilon}\left[\log \frac{p_\eta(\mathcal{G}_t \mid \mathcal{P})}{p_\theta(\mathcal{G}_t \mid \mathcal{P})}- \omega r(\mathcal{G}_t, \mathcal{P})\right] \\ & =\min _{\pi} \mathbb{E}_{\mathcal{P}} \mathbb{E}_{\mathcal{G}_t,\epsilon} \\
    & \left[\log \frac{1}{\frac{1}{Z(\mathcal{G}_t)} p_\theta(\mathcal{G}_t \mid \mathcal{P}) \exp \left(\omega r(\mathcal{G}_t, \mathcal{P})\right)}-\log Z(\mathcal{G}_t)\right]
\end{aligned}
\label{eq:SM_derivation}
\end{equation}
where 
\begin{equation}
    Z(\mathcal{P}) = \mathbb{E}_{\mathcal{G}_0,t,\epsilon} p_{\theta,t}(\mathcal{G}_t|\mathcal{P}) \exp(\omega r(\mathcal{G}_t,\mathcal{P}))
\end{equation}

Note that the partition function is a function of only $\mathcal{G}_t$ and the pre-training distribution $p_\theta$, but does not depend on the generative distribution $p_\eta$. We can now define
\begin{equation}
     p_{\eta,t}^*(\mathcal{G}_t|\mathcal{P})= \frac{p_{\theta,t}(\mathcal{G}_t|\mathcal{P})) \exp(\omega r(\mathcal{G}_t,\mathcal{P}))}{Z(\mathcal{P})}
\label{eq:SM_global_optimal}
\end{equation}

Since $p_{\eta,t}^*(\mathcal{G}_t|\mathcal{P}) \geq 0$, $\sum_\mathcal{P} p_{\eta,t}^*(\mathcal{G}_t|\mathcal{P}) = 1$ and $Z(\mathcal{P})$ is not related to $\mathcal{G}_t$, the final objective in~\cref{eq:SM_derivation} can be rewritten as
\begin{equation}
    \begin{array}{c}
    \min _{\eta} \mathbb{E}_{\mathcal{P}}\left[\mathbb{E}_{\mathcal{G}_t,t}\left[\log \frac{p_\eta(\mathcal{G}_t \mid \mathcal{P})}{p_\eta^{*}(\mathcal{G}_t \mid \mathcal{P})}\right]-\log Z(\mathcal{P})\right]= \\ 
    \min _{\eta} \mathbb{E}_{\mathcal{P}}\left[\mathbb{D}_{\mathrm{KL}}\left(p_\eta(\mathcal{G}_t \mid \mathcal{P}) \| p_\eta^{*}(\mathcal{G}_t \mid \mathcal{P})\right)-\log Z(\mathcal{P})\right]
    \end{array}
\end{equation}
According to Gibbs’ inequality, the KL divergence is minimized if and only if the two distributions are the same. Thus, the optimal solution is
\begin{equation}
     p_{\eta,t}(\mathcal{G}_t|\mathcal{P})=p_{\eta,t}^*(\mathcal{G}_t|\mathcal{P})= \frac{p_{\theta,t}(\mathcal{G}_t|\mathcal{P})) \exp(\omega r(\mathcal{G}_t,\mathcal{P}))}{Z(\mathcal{P})}
\end{equation}

\section{Additional completion scenes}
\begin{figure*}
    \centering
    \includegraphics[width=0.8\linewidth]{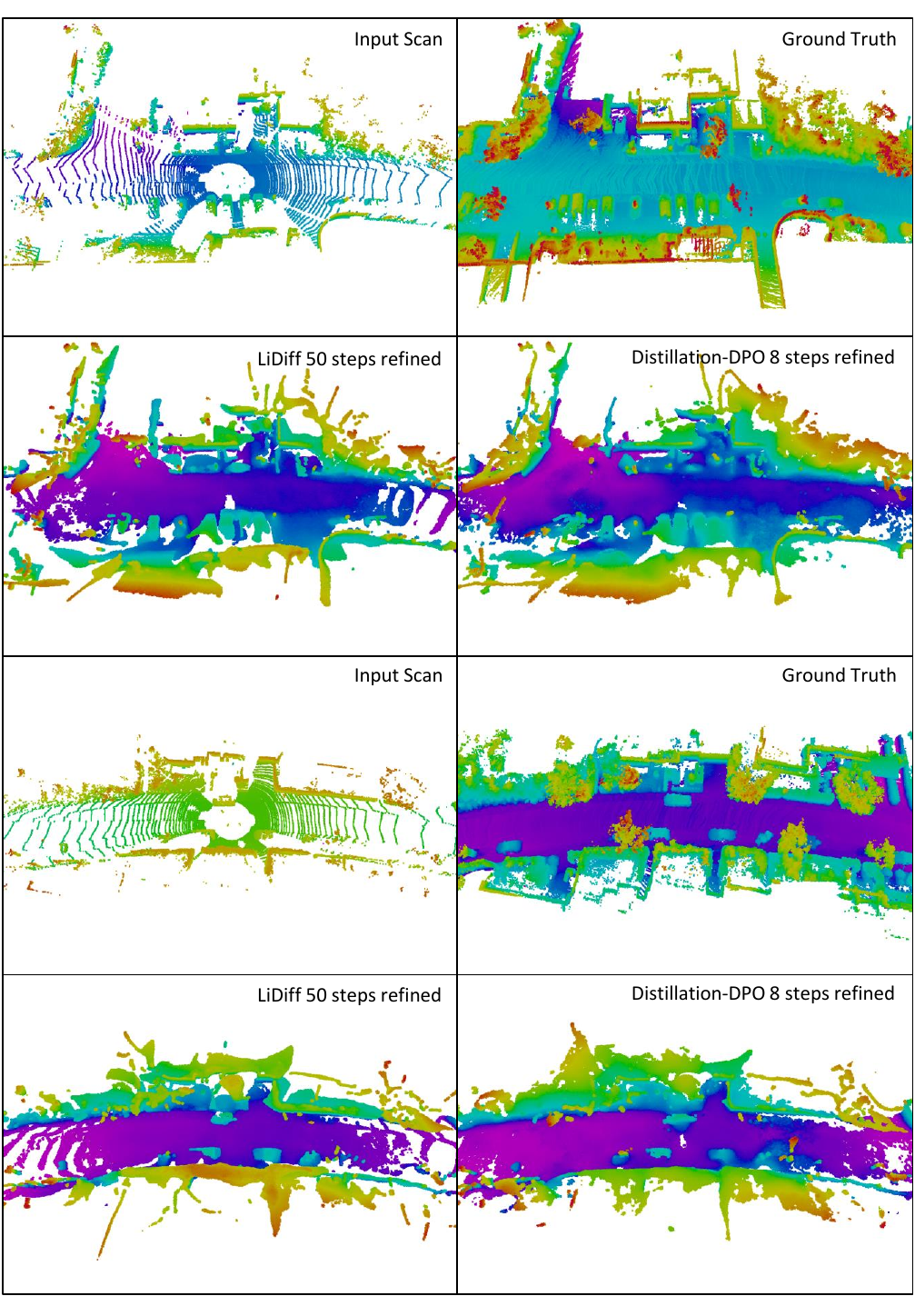}
    \caption{Additional completed samples of Distillation-DPO from SemanticKITTI dataset.}
    \label{fig:addition1}
\end{figure*}

\begin{figure*}
    \centering
    \includegraphics[width=0.8\linewidth]{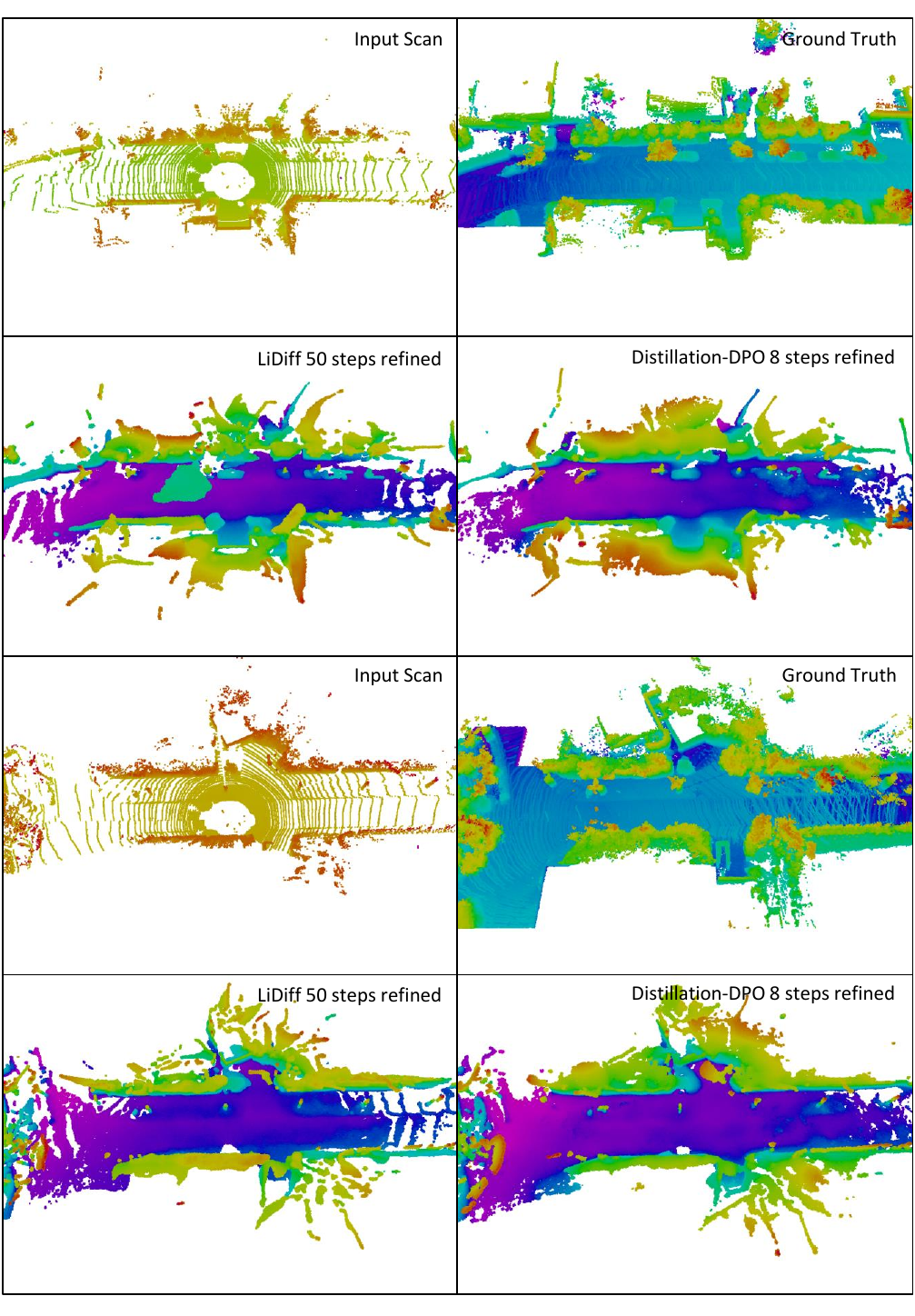}
    \caption{Additional completed samples of Distillation-DPO from SemanticKITTI dataset.}
    \label{fig:addition2}
\end{figure*}

\begin{figure*}
    \centering
    \includegraphics[width=0.8\linewidth]{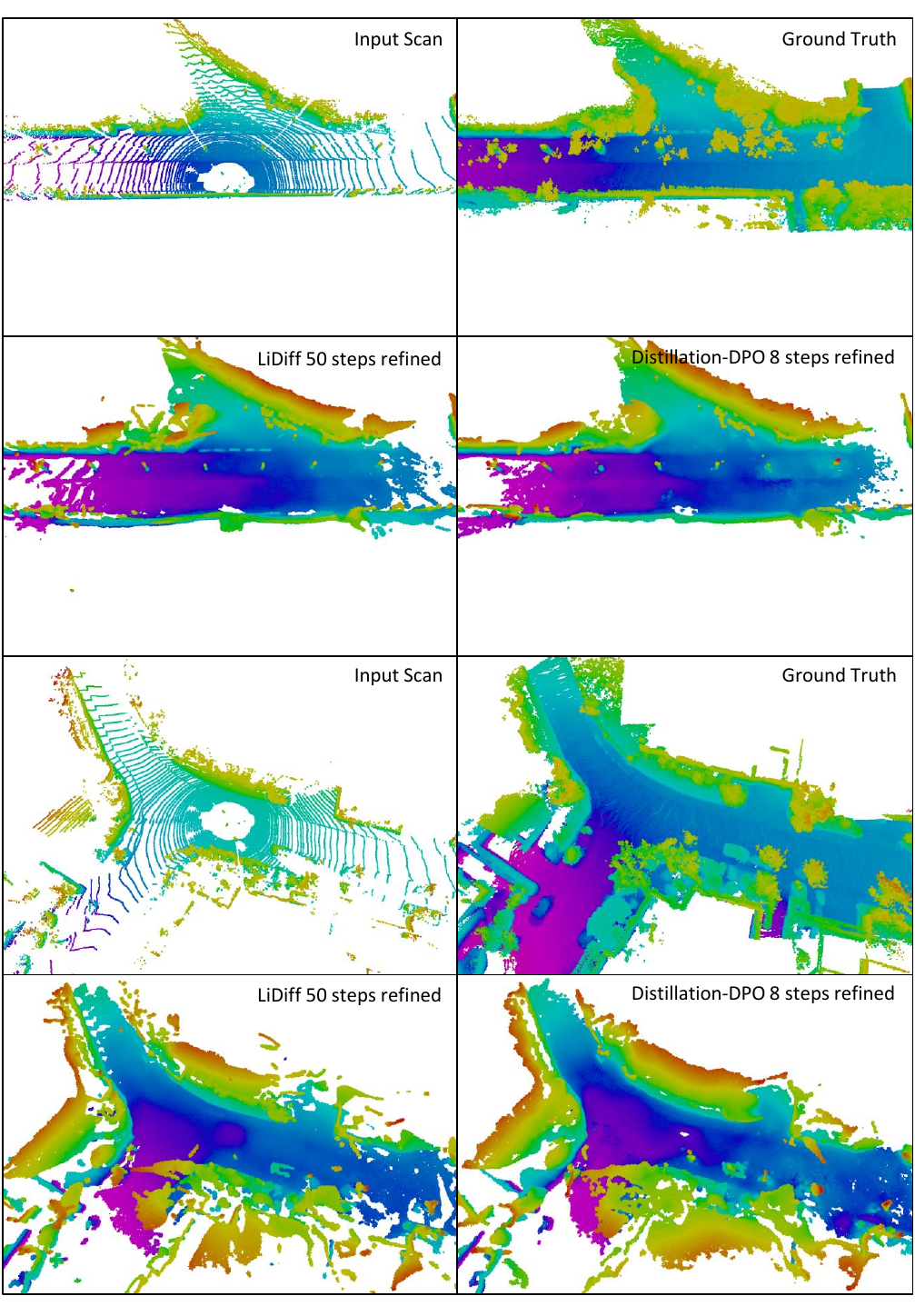}
    \caption{Additional completed samples of Distillation-DPO from SemanticKITTI dataset.}
    \label{fig:addition3}
\end{figure*}

\cref{fig:addition1}, \cref{fig:addition2} and \cref{fig:addition3} show more completed scene by Distillation-DPO on SemanticKITTI dataset. Compared with the 50-step completion results of LiDiff~\cite{LiDiff}, Distillation-DPO can achieve higher completion quality in a significantly shorter time, with improved scene completeness and clearer local details, such as vehicles and other objects.

\section{More experiment results}
\subsection{Ablation study on scene occupancy}
To further evaluate the quality of the completed scenes, we compute the Intersection-over-Union (IoU)~\cite{iou} between the completed scene and the ground truth. IoU measures the occupancy consistency by calculating the overlap between the voxels in the completed scene and the ground truth. A higher IoU value indicates a greater overlap, meaning the completed scene more closely matches the real scene. The IoU results vary based on different voxel resolutions, as lower resolutions capture finer-grained details during computation. Here, we consider three voxel resolutions: $0.5m$, $0.2m$, and $0.1m$.

However, as a voxel-based metric, IoU is effective in evaluating LiDAR scene completion quality in signed distance field (SDF)-based methods. In contrast, Distillation-DPO is a point cloud-based completion method, making IoU prone to bias when assessing completion quality. Therefore, we only provide IoU results as a reference. As shown in~\ref{tab:occupancy1}, Distillation-DPO achieves IoU values comparable to existing models, especially when the voxel resolution is smaller. Thus, Distillation-DPO can achieve completion results similar to or even better than LiDiff while maintaining faster completion speed, further demonstrating its effectiveness.

\begin{table}[t]
\centering
\resizebox{\columnwidth}{!}{%
\begin{tabular}{lrrc}
\toprule
\multirow{2}{*}{Model} & \multicolumn{3}{c}{SemanticKITTI (IoU) \% $\uparrow$} \\ \cline{2-4} 
 & \multicolumn{1}{c}{0.5m} & \multicolumn{1}{c}{0.2m} & 0.1m \\ \midrule
LMSCNet~\cite{lmscnet} & 32.23 & 23.05 & 3.48 \\
LODE~\cite{lode} & \cellcolor[RGB]{\colorsecond}{43.56} & \cellcolor[RGB]{\colorfirst}{47.88} & 6.06 \\
MID~\cite{MID} & \cellcolor[RGB]{\colorfirst}{45.02} & \cellcolor[RGB]{\colorsecond}{41.01} & \cellcolor[RGB]{\colorthird}{16.98} \\
PVD~\cite{pvd} & 21.20 & 7.96 & 1.44 \\
LiDiff~\cite{LiDiff} & \cellcolor[RGB]{\colorthird}{42.49} & 33.12 & 11.02 \\
LiDiff (Refined)~\cite{LiDiff} & 40.71 & \cellcolor[RGB]{\colorthird}{38.92} & \cellcolor[RGB]{\colorfirst}{24.75} \\ \midrule
Distillation-DPO & 38.24 & 30.05 & 11.15 \\
Distillation-DPO (Refined) & 37.36 & 31.61 & \cellcolor[RGB]{\colorsecond}18.74 \\ \bottomrule
\end{tabular}%
}
\caption{The IoU evaluation results on the SemanticKITTI dataset.}
\label{tab:occupancy1}
\end{table}

\subsection{Experiments on different training strategy}
We also compare different training strategies. In this work, the student model is trained using single-step generation for completion scenes, while 8-step sampling is used during evaluation. To analyze the impact of sampling steps during training, we conduct experiments where the student model performs random 1-8 step sampling during training. The results in Tab~\ref{tab:strategy} show that when the student model performs random 1-8 step sampling, Distillation-DPO’s performance noticeably declines, which demonstrates the effectiveness of the proposed training strategy in Distillation-DPO.

\begin{table}[t]
\centering
\resizebox{\columnwidth}{!}{%
\begin{tabular}{lrrr}
\toprule
Model & \multicolumn{1}{c}{CD $\downarrow$} & \multicolumn{1}{c}{JSD $\downarrow$} & \multicolumn{1}{c}{EMD $\downarrow$}  \\ \midrule
Distillation-DPO & \cellcolor[RGB]{\colorfirst}0.354 & \cellcolor[RGB]{\colorfirst}0.387 & \cellcolor[RGB]{\colorfirst}23.66 \\
Distillation-DPO (Random inference step) & 0.621 & 0.533 & 28.92 \\ \bottomrule
\end{tabular}%
}
\caption{Comparison results of using different teacher model. LiDiff$^*$ represents the LiDiff model refined with Diffusion-DPO. Distillation-DPO$^*$ represents the Distillation-DPO trained with LiDiff$^*$. All results have been refined.}
\label{tab:strategy}
\end{table}

\subsection{Experiment on the preference data pairs}
\begin{figure}
    \centering
    \includegraphics[width=\columnwidth]{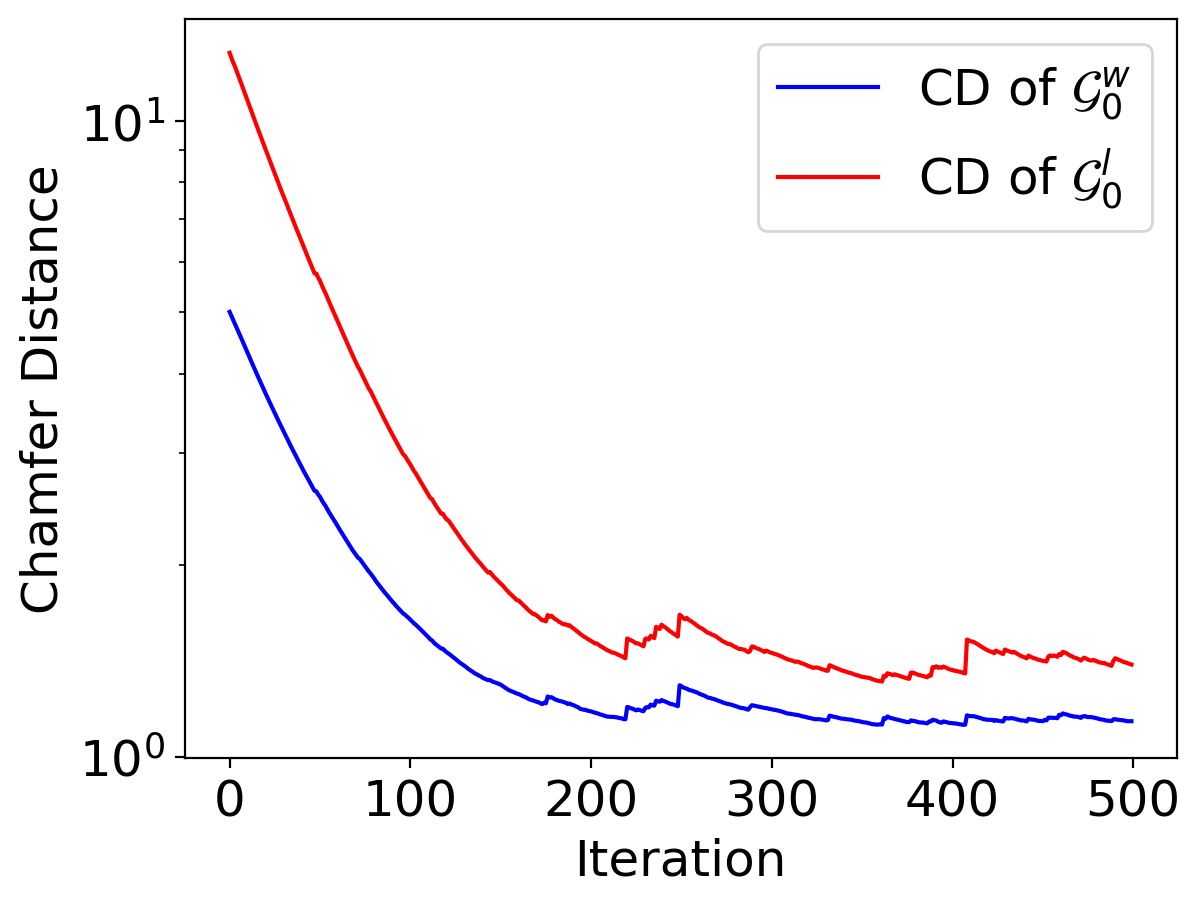}
    \caption{Additional completed samples of Distillation-DPO from SemanticKITTI dataset.}
    \label{fig:cd_chart}
\end{figure}

In this part, we visualized the Chamfer Distance values of the winning samples $\mathcal{G}_0^w$ and losing samples $\mathcal{G}_0^l$ during training, as shown in~\cref{fig:cd_chart}. It can be observed that the Chamfer Distance values of $\mathcal{G}_0^w$ consistently remain lower than those of $\mathcal{G}_0^l$ and decrease as training progresses. This demonstrates that the proposed method of using Chamfer Distance to select preference data is effective.




    



\section{Failure examples}
\begin{figure*}
    \centering
    \includegraphics[width=0.8\linewidth]{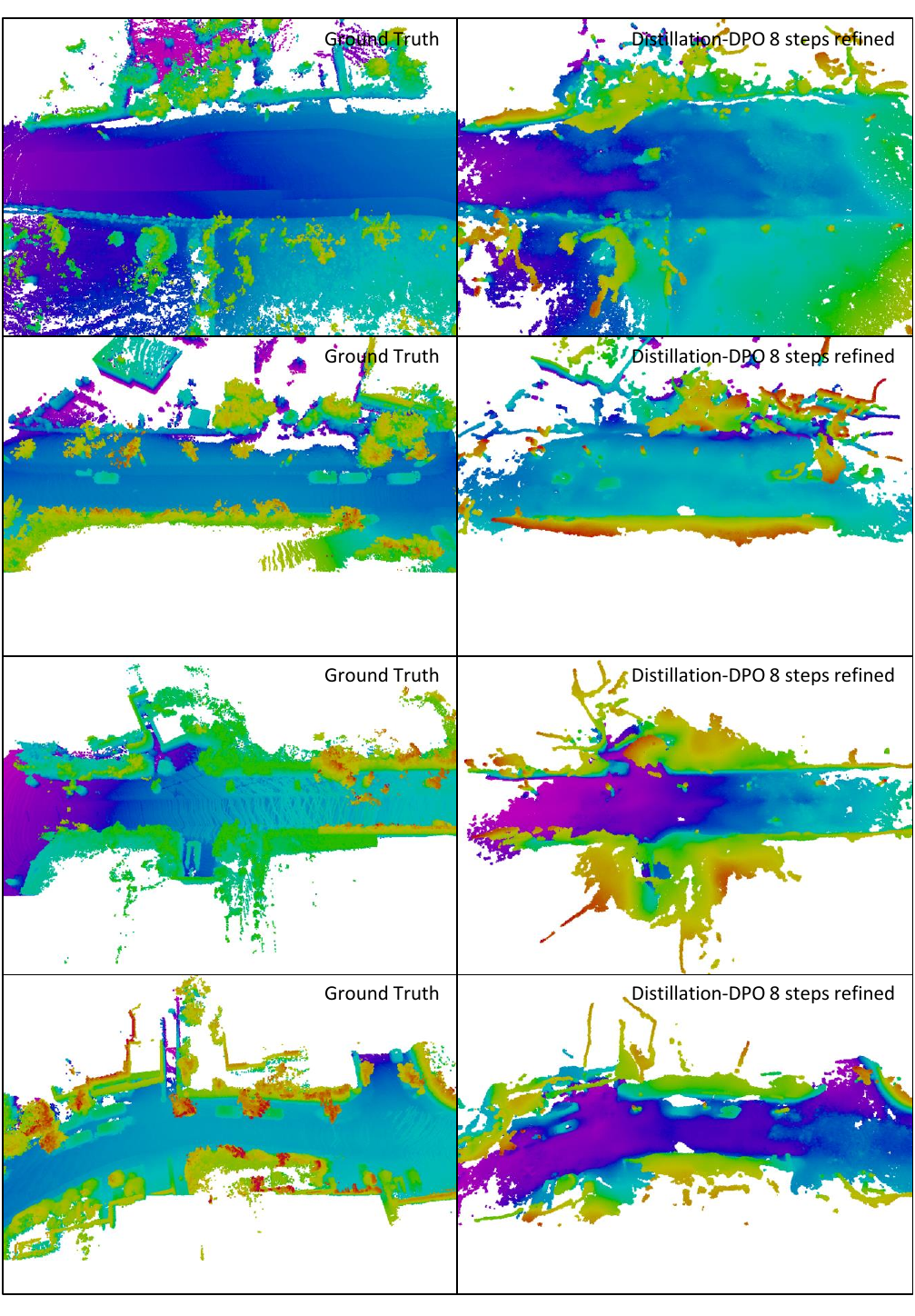}
    \caption{Failure examples of Distillation-DPO.}
    \label{fig:failure}
\end{figure*}

\cref{fig:failure} presents some failure cases of Distillation-DPO. The occurrence of these failure cases may be due to the limitations inherent in Distillation-DPO. First, during scene completion, the points in sparse scan $\mathcal{P}$ need to be duplicated $K$ times to serve as the initial dense point cloud. However, the number of points in different real-world scenes varies. When the number of points in the initial dense point cloud is smaller than that in the ground truth, certain parts of the scene may be missing. Conversely, when the number of points exceeds that of the ground truth, noise artifacts or overly smooth surfaces may appear. Second, the performance of Distillation-DPO is limited by the performance of the teacher model. Although the student model in this paper ultimately outperforms the teacher model, it still retains some characteristics of the teacher model. As a result, issues present in the teacher model, such as incomplete scene completion, may also appear in the student model’s completions.

{
    \small
    \bibliographystyle{ieeenat_fullname}
    \bibliography{main}
}

